\documentclass[letterpaper, 10 pt, conference]{ieeeconf}  

\usepackage{graphicx}
\usepackage{units} 
\usepackage{amsmath} 

\IEEEoverridecommandlockouts                              

\title{\LARGE \bf
Real-Time Visual Localisation in a Tagged Environment
}

\author{J\'er\'emy Taquet$^{1}$, Ga\"el \'{E}corchard$^{1}$, and Libor P\v{r}eu\v{c}il$^{1}$
\thanks{$^{1}$Authors are with the Czech Institute of Informatics, Robotics and Cybernetics, Czech Technical University in Prague, Czech Republic,
        {\tt\small jeremy.taquet@sigma-clermont.fr}}%
}

\begin{document}

\maketitle
\thispagestyle{empty}
\pagestyle{empty}

\begin{abstract}

In a robotised warehouse a major issue is the safety of human operators in case of intervention in the work area of the robots.
The current solution is to shut down every robot but it causes a loss of productivity, especially for large robotised warehouses.
In order to avoid this loss we need to ensure the operator's security during his/her intervention in the warehouse without powering off the robots.
The human operator needs to be localised in the warehouse and the trajectories of the robots have to be modified so that they do not interfere with the human.
The purpose of this paper is to demonstrate a visual localisation method with visual elements that are already available in the current warehouse setup.

\end{abstract}

\section{Introduction}

In order to gain productivity and reduce costs of operation, many warehouse equipment suppliers propose now fully robotised solutions.
The context of the presented research are warehouse where goods for the end-user or products in the Business-to-business sector are stored, commissioned and shipped, which gains relevance with the growing e-commerce market.
A method for the management of such a robotised warehouse is to store items on shelves that are moved around by robots between storage space and a so-called pick station, where a human picks single items from the shelf and puts them directly into the box that will be send to the end-customer.
In such a scenario, the robots are confined in the storage space.
For the maintenance of such systems or for other special actions, such as tidying a dropped item, human intervention is often required.
The current solution is to shut down the complete automatised system as soon as a human operator enters the protected area.
Every intervention is then costly because all robots and not only the faulty one stop to be productive.
In order to allow a human to collaborate with the robots, his/her position has to be known by the robot manager.
For the localisation to be as cost-effective as possible, the use of available information or devices is preferred.
This is the reason why a visual-based localisation system associated with available ground markers was chosen and will be presented here.
It is to be noted that this localisation system does not have any safety issue because the safety will be realised by other components.

\section{Existing Solutions}

Today there already exist solutions to localise a camera in an unknown environment for example with SLAM (Simultaneous Localisation And Mapping) algorithms \cite{SLAM} but it will take too much computation time for our usage.
This kind of localisation already exists for other known environment too but to do it in real-time with a nanocomputer we must improve the computation time.
The solution must find the position of the operator while he is moving.

\section{Methods and Materials}

\subsection{Context Definition}

\begin{figure}[h]
  \centering
  \includegraphics[width=5cm]{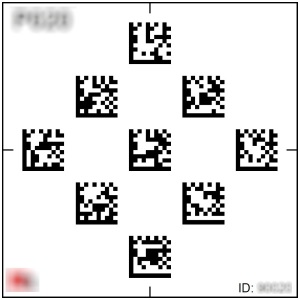}
  \caption{Example of a ground sticker}
  \label{fig:pattern}
\end{figure}

For maintenance tasks or special tasks that the highly specialized robots cannot accomplish, a human operator evolves in a warehouse that can be relatively dark (between 50 and 200 lux on the ground) and where robots are moving, either carrying a shelf or not.
The main goal of this localisation is to send the position of the operator to the robots' management system in real-time to have the robots avoid to go near the human for his/her own safety.
For the communication with the warehouse system and for the localisation system, the operator will wear a safety vest equipped with a computer and one or more cameras to compute his position.
The localisation system should then be portable, without delay, and preferably on board to reduce the communication with a remote computer, difficult in a environment full of obstacles for radio communication.
Another constraint on the localisation system is that it should be scalable without significant impact on the warehouse cost.
For the warehouse operation some stickers are located on the ground at known positions, spaced more or less regularly with a distance of less than two meters between two stickers.
These stickers were chosen as the items to be recognised by our localisation algorithm.
Each sticker, shown in Fig.~\ref{fig:pattern}, comprises of a set of Data Matrix codes, which allows to identify it uniquely, enclosed in a 10-centimeter wide square.

\subsection{Presentation of the Solution}

To calculate the position of the human, we first have to detect the visual landmarks, the ground stickers.
By reading one of the sticker's Data Matrix codes we know which pattern has been read so we know the exact position of the detected pattern in the warehouse.
To develop this solution some open-source libraries were used (OpenCV \cite{OpenCV} and libdmtx \cite{libdmtx}) in different algorithms and programs.

With the knowledge of the absolute position of the sticker in the warehouse, we can compute the position of the camera with a correctly parameterised pinhole model, and such, the absolute position of the human operator.

\subsection{Detailed Description of the Localisation System}

The algorithm for the detection of the visual landmarks and Data Matrix decoding is divided into four parts: 
\begin{itemize}
  \item detection of the stickers in the image;
  \item extraction of a Region of Interest (ROI) around each sticker;
  \item reading of Data Matrix codes in the ROI;
  \item computing of the camera's absolute position according to the position of the sticker in the warehouse and the position of its projection in the image.
\end{itemize}

\subsubsection{Pattern Recognition}

\begin{figure}[!ht]
  \centering
  \includegraphics[width=8.5cm]{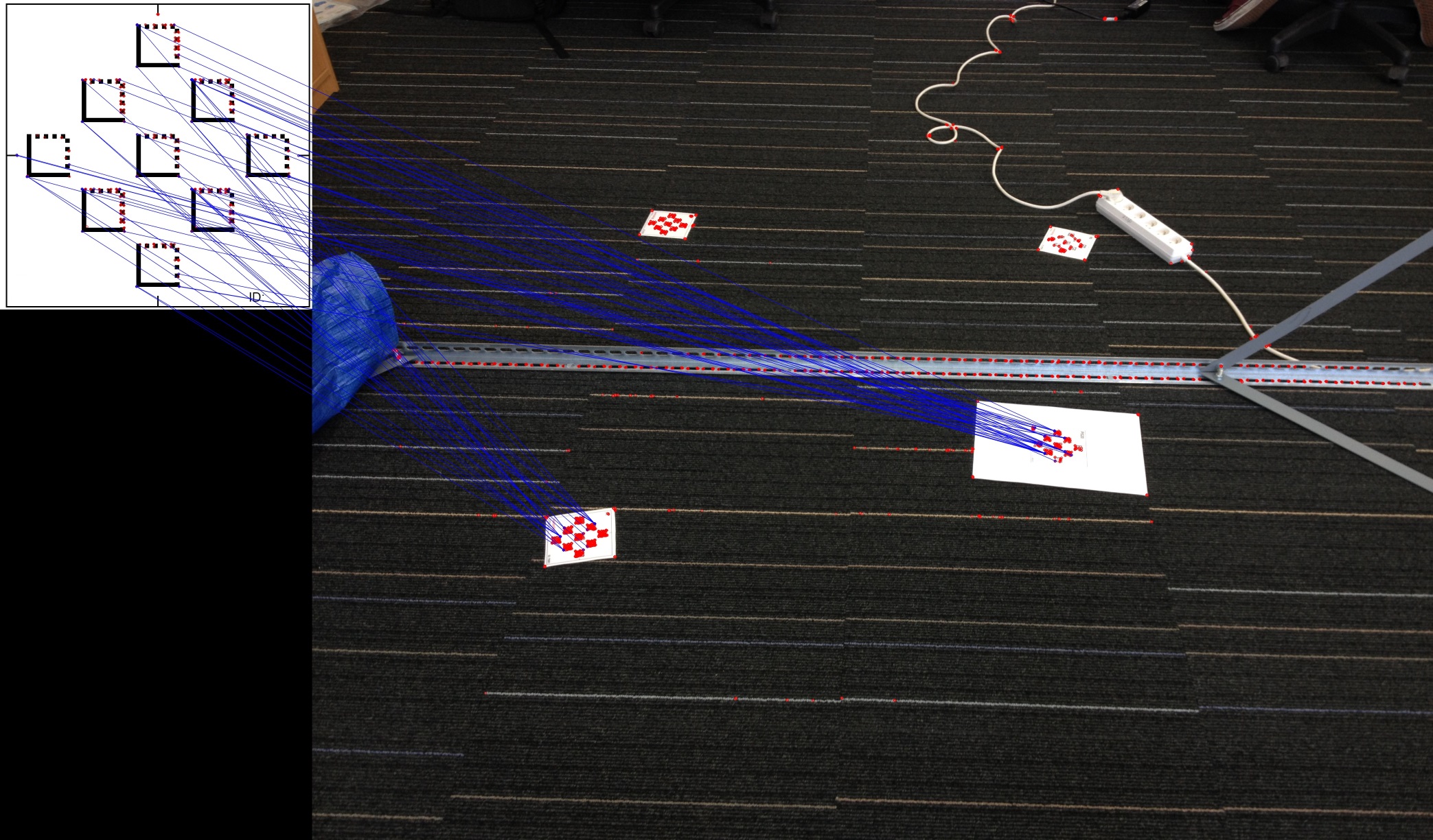}
  \caption{ORB feature detector}
  \label{fig:orb}
\end{figure}

As a first step, we need to detect the pattern in the current frame of the video stream.
The method is based on feature matching.
The ORB features were chosen because of their rotation invariance.
Indeed, it is not known in advance from which side of the sticker the operator will approach the stickers and the sticker can have any orientation in the image.
At first, we load a reference picture, a sticker with random Data Matrix codes with the same coding (10$\times$10).
Then, we convert the picture into greyscale and use detectors and ORB descriptors \cite{Rublee11a} to characterize the picture off-line.
The size of the sticker in the reference image should correspond to what will be expected during the operation in the warehouse to avoid scaling problems.
After that, during normal warehouse operation, we apply the same treatment to the live picture stream captured by the camera and use the nearest neighbour search method to match the features between the reference picture and the scene picture.
If we have more than 50 matches we considered that a sticker is detected in the picture, as we can see in Fig.~\ref{fig:orb}.
Without sticker in the scene we usually have less than 15 matches.
On Fig.~\ref{fig:orb} we don't detect the 2 stickers in the background because of their perspection deformation.

\subsubsection{Extraction of ROI Around Stickers}
\label{sec:ROI}

\begin{figure}[!ht]
  \centering
  \includegraphics[width=8.5cm]{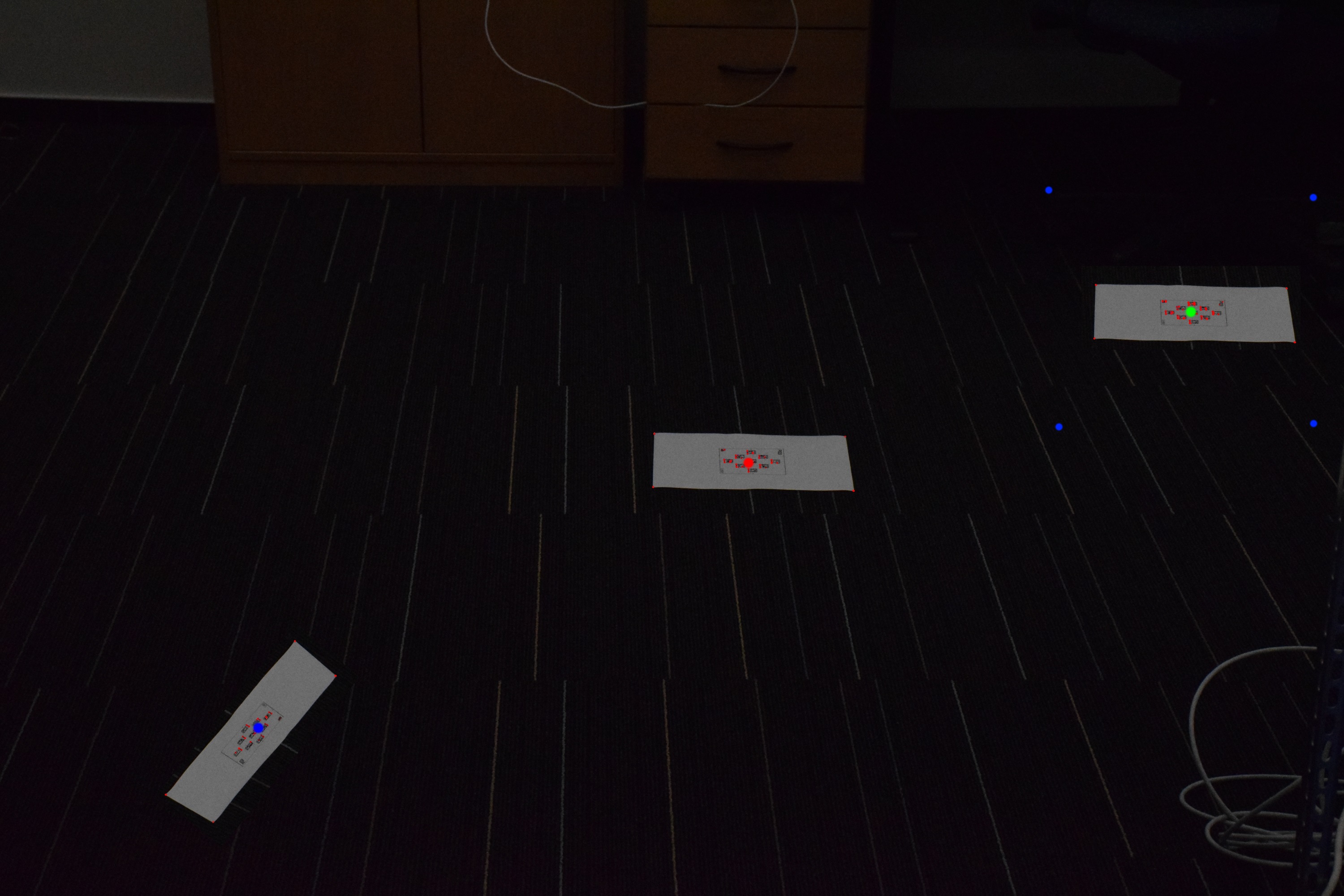}
  \caption{K-means clustering ans ROI definition}
  \label{fig:clustering}
\end{figure}

After having ensured that at least one sticker is detected in the current image, we must define an ROI around each sticker.
The first aim of these ROIs is to reduce the computation time of the decoding algorithm.
The second aim is to be able to distinguish between the stickers because the decoding algorithm, libdmtx, does not return the position in the image of the decoded Data Matrix codes.
We detect these ROI with a K-means algorithm, \cite{MacQueen67a}.
Our K-mean algorithm start with 4 clusters and merges clusters if their means are too close, to have, at the end, one cluster per sticker (we cannot have more than 3 sticker in the picture according to their position in the warehouse).
We cluster features in different clouds of points to choose the cloud with the most points and cut around the selected cloud of points.
We need only one sticker so we ignore clusters with lower count of features.
A test of clustering and ROI definition is presented in Fig.~\ref{fig:clustering}, on this figure we can see many detected points on each sticker, 3 points as the mean of each cluster in blue, green and red and the ROI materialised with 4 blue points around the upper sticker.

\subsubsection{Data Matrix Decoding}

Once stickers are detected and isolated, we must read the Data Matrix code in it.
We tested two variants for the Data Matrix decoding.
In the first one, the corner of the square surrounding the Data Matrix codes are detected and the perspective is corrected with OpenCV's 
warpPerspective function.
The ROI with corrected perspective is then sent to libdmtx.
In the second variant, the ROI is given as is over to libdmtx.
Both variants gave similar results in term of speed and accuracy but we chose the second one.
The localisation system then communicates with the warehouse management system and receives the absolute position of the pattern stored in a database.

\subsubsection{Position Computation}

\begin{figure}[!ht]
  \centering
  \includegraphics[width=8.5cm]{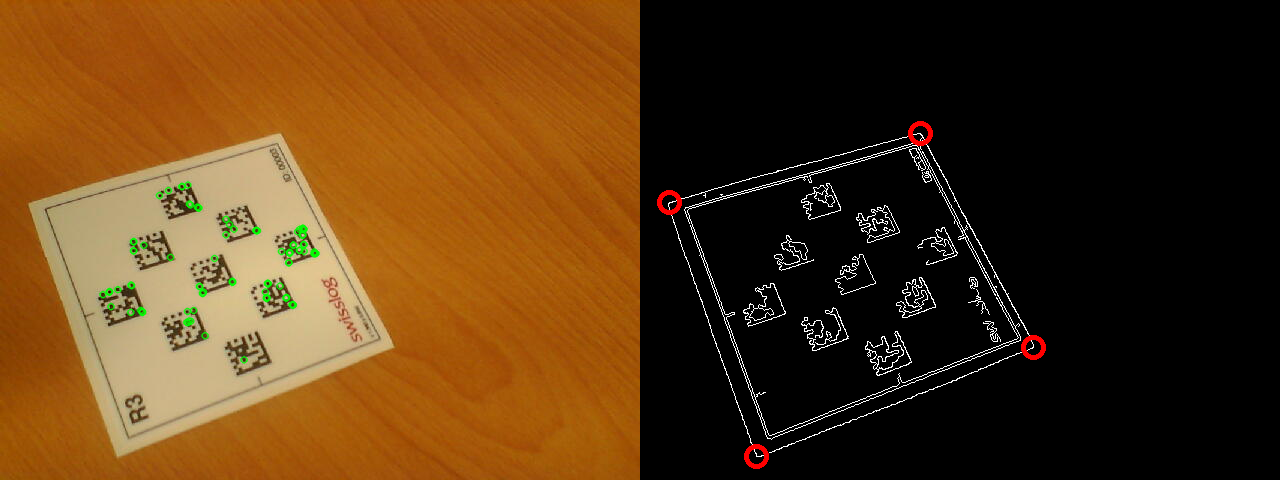}
  \caption{contour and corners detection}
  \label{fig:corners}
\end{figure}

Finally, after knowing the absolute position of the sticker we can compute the position of the camera according to the sticker.
We start by detecting the contour of the pattern and extracting every points of this contour to get its corners.
The corner extraction is done easily by looking at the extreme points of this contour as it's done in the Fig.~\ref{fig:corners}.
To compute the position of the sticker according to its position in the picture, we use an OpenCV's function solvePnp so we need at least 3 points to compute its position that's why we use the corners of the sticker.

We can compute the position of the camera with the solvePnp function (pinhole model) using the 4 corners, the intrinsic parameter's matrix of the camera and the absolute position of the sticker in the warehouse.

The pinhole model is given by
\begin{equation}
  \begin{pmatrix}
    su \\
    sv \\
    s
  \end{pmatrix}
  =
  \mathbf{K}
  \mathbf{F}
  \mathbf{T}
  \begin{pmatrix}
    X \\
    Y \\
    Z \\
    1
  \end{pmatrix}
\end{equation}
with
$\mathbf{K}
= 
\begin{pmatrix}
  k_u & s_{uv} & c_u \\
  0 & k_v & c_v \\
  0 & 0 & 1
\end{pmatrix}
$,
$\mathbf{F}
=
\begin{pmatrix}
  f & 0 & 0 & 0 \\
  0 & f & 0 & 0 \\
  0 & 0 & 1 & 0
\end{pmatrix}
$,
$
\mathbf{T}
=
\begin{pmatrix}
   &         &   & t_x \\
   & R_{3\times 3} &   & t_y \\
   &         &   & t_z \\
 0 & 0       & 0 & 1
\end{pmatrix}
$,\\\\

$  \begin{pmatrix}
    X \\
    Y \\
    Z \\
    1
  \end{pmatrix}$ are the coordinates of a point $M$ in space,\\

$  \begin{pmatrix}
    su \\
    sv \\
    s
  \end{pmatrix}$ are the coordinates of a the image of the point $M$ in the picture,\\
$k_u$ and $k_v$ the magnification factors,\\
$c_u$ and $c_v$ the coordinates of the projection of the camera's optical centre on the image plane,\\
$s_uv$ represent the non-orthogonality of lines and columns of camera's photosensitive cells,\\
$f$ the focal length,\\
$R_{3\times 3}$ a rotation matrix in Euclidean space,\\
$(t_x, t_y, t_z)$ a translation vector.\\

\subsubsection{Alternative to Data Matrix Decoding}

After the first trials of the algorithm we noticed that we were not able to read the Data Matrix codes on most of the frames because of the motion-blur (the camera is ''held'' by a walking human).
To avoid this blur we have to increase the shutter speed.
To have a sharp picture the projection of the objects in the image frame must not move more than one pixel.
To compute the displacement of the projections of objects in the image frame, we have: 
\begin{equation}
  R = \frac{f}{D}
\end{equation}
and
\begin{equation}
  d = R \frac{V}{N}
\end{equation}
with:\\
$R:$ magnification factor,\\
$f:$ focal length (\unit{m}),\\
$D:$ distance between camera and the object (\unit[1]{m} here),\\
$V:$ velocity of the camera movement (\unit[1]{m/s} here),\\
$\frac{1}{N}:$ shutter speed (\unit{s$^{-1}$}),\\
$d:$ distance travelled by the projection of the object in the picture frame (\unit{m}).\\

\begin{figure}[!ht]
  \centering
  \includegraphics[width=8.5cm]{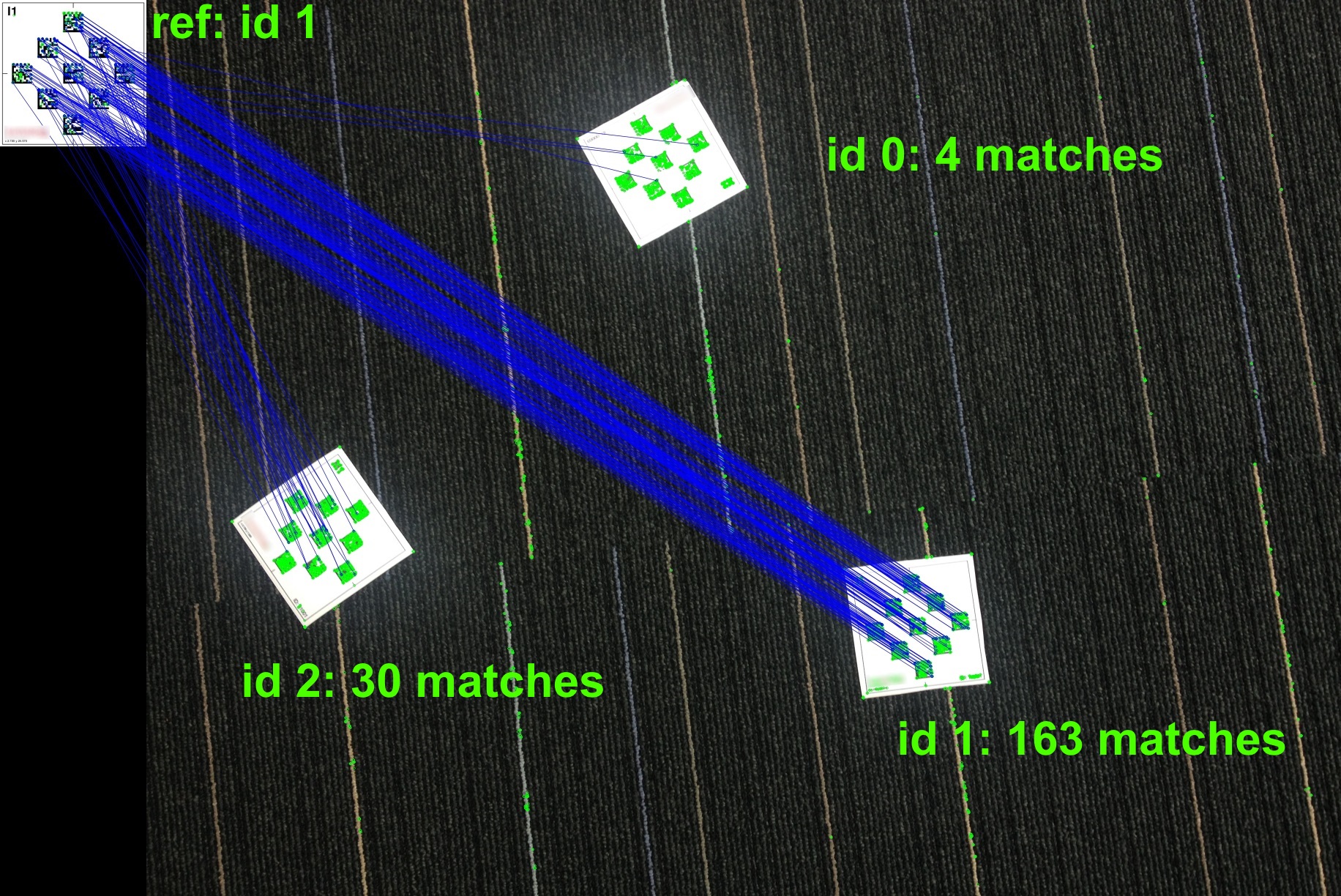}
  \caption{recognition of pattern with reference id1}
  \label{fig:multi-ref1}
\end{figure}

\begin{figure}[!ht]
  \centering
  \includegraphics[width=8.5cm]{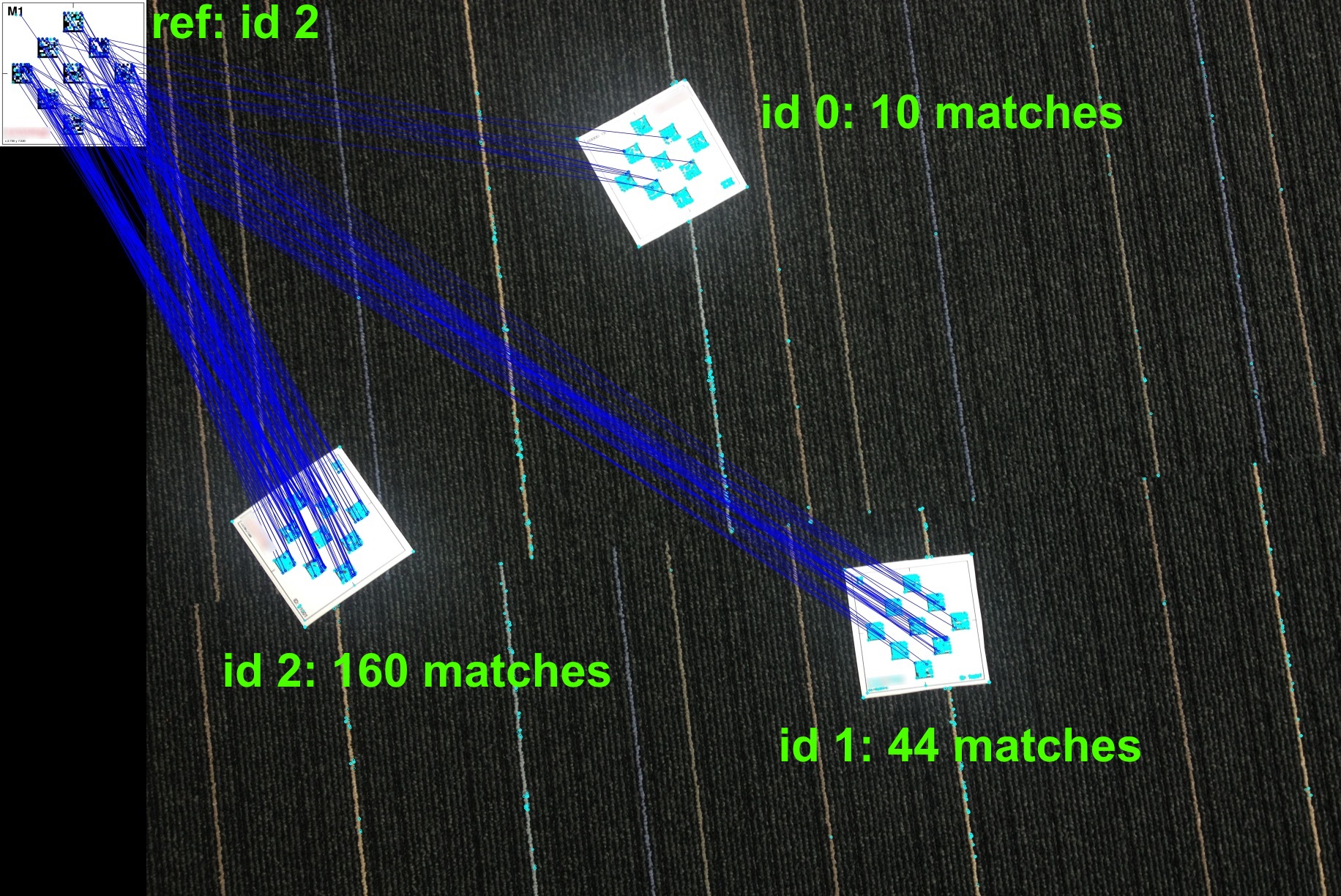}
  \caption{recognition of pattern with reference id2}
  \label{fig:multi-ref2}
\end{figure}

For our experiments, we used a Raspberry Pi and the official \unit[5]{MP} Raspberry Pi camera with a pixel size of \unit[1.4]{$\mu$m$^2$} and a focal length of \unit[3.6]{mm}
We would thus need a shutter speed of $\frac{1}{N} < \unit[2571]{\mathrm{s}^{-1}}$ to reach this constraint.
Such shutter speed were not reachable because the quantity of light hitting the sensor would be too small.
Because of this problem, libdmtx is not able to read Data Matrix codes every time, so we can try to recognize pattern without reading these codes.
In that purpose we save each sticker as references in the embedded computer, we have to compute the ORB descriptors on each reference.
This computation is advantageously done off-line.
Then we compute ORB descriptors in the picture stream.
We use the same algorithm as in Section~\ref{sec:ROI} but with a much larger count of features.
This allows us to identify the sticker on the ground as being the one with the highest count of matches with its pre-computed reference, this replaces the Data Matrix decoding.
As can be seen on Fig.~\ref{fig:multi-ref1} and~\ref{fig:multi-ref2}, the algorithm detects all stickers but we have much more correspondences with the sticker corresponding to the reference.
On these pictures we have 500 features on each reference, 10000 features for the scene picture and we have around 160 matching features for the good sticker and between 4 and 50 matching features for the wrong stickers.
This test was done with pictures so not in real-time, with this amount of features to match, we are not able to do it in real-time with the current setup.
To reach real-time treatment it needs a higher computation power to match the features, we have to be able to compare many references with the same scene picture which is power consuming.
With an estimation of the position, computed with last know position, we can improve the efficiency of our algorithm by comparing the scene picture only with references  that have reasonable probability to be detected.

\section{Results and Discussion}

After many tests, the detection of pattern work correctly  in the test environment, we have some problem with the reading of Data matrix, so we frequently miss a landmark (the algorithm recognise the sticker but it is not able to read the Data matrix).
The next step is to test the algorithm in the real environment to adapt parameters.
We will also have to choose a new camera better adapted to the low light environment and high shutter speed requirement.


\addtolength{\textheight}{-12cm}   




\section*{ACKNOWLEDGEMENT}

This work was developed within the SafeLog project funded by the European Union's Horizon 2020 research and innovation programme under grant agreement No 688117.


\bibliography{bibliography}

\begin{thebibliography}{1}

\bibitem{OpenCV}
Open source computer vision library.
\newblock https://opencv.org.

\bibitem{SLAM}
Andrew~J. Davison.
\newblock Real-time simultaneous localisation and mapping with a single camera.
\newblock {\em Proceedings of the Ninth IEEE International Conference on
  Computer Vision}, 2003.

\bibitem{libdmtx}
Mike Laughton.
\newblock Open source data matrix software \& library.
\newblock https://libdmtx.sourceforge.net.

\bibitem{MacQueen67a}
J.~MacQueen.
\newblock Some methods for classification and analysis of multivariate
  observations.
\newblock In {\em Proceedings of the Fifth Berkeley Symposium on Mathematical
  Statistics and Probability}, volume~1, pages 281--297. The Regents of the
  University of California, 1967.

\bibitem{Rublee11a}
E.~Rublee, V.~Rabaud, K.~Konolige, and G.~Bradski.
\newblock {ORB}: An efficient alternative to {SIFT} or {SURF}.
\newblock In {\em International Conference on Computer Vision}, pages
  2564--2571, November 2011.
\newblock ICCV 2011.

\end{thebibliography}
\bibliographystyle{plain}

\end{document}